# Rotational Augmentation Techniques: A New Perspective on Ensemble Learning for Image Classification

Unai Muñoz-Aseguinolaza[1], Basilio Sierra[1] and Naiara Aginako[1]

[1] Department of Computer Science and Artificial Intelligence, University of the Basque Country, Donostia-San Sebastián, 20018, Gipuzkoa, Spain
unai.munoza@ehu.eus, b.sierra@ehu.eus, naiara.aginako@ehu.eus

## ABSTRACT

*The popularity of data augmentation techniques in machine learning has increased in recent years, as they enable the creation of new samples from existing datasets. Rotational augmentation, in particular, has shown great promise by revolving images and utilising them as additional data points for training. This research study introduces a new approach to enhance the performance of classification methods where the testing sets were generated employing transformations on every image from the original dataset. Subsequently, ensemble-based systems were implemented to determine the most reliable outcome in each subset acquired from the augmentation phase to get a final prediction for every original image. The findings of this study suggest that rotational augmentation techniques can significantly improve the accuracy of standard classification models; and the selection of a voting scheme can considerably impact the model's performance. Overall, the study found that using an ensemble-based voting system produced more accurate results than simple voting.*

## 1. INTRODUCTION

Data mining is the process of applying statistical and machine learning (ML) procedures to extract knowledge and hidden relationships from large datasets. The main goal is to transform data into practical information that can be used for decision-making, such as predicting future outcomes and optimising processes. Image classification is one of the many techniques used in data mining to build predictive models, which involves categorising an image based on its characteristics.

Feature extraction and attribute selection are essential techniques used in classification tasks to improve the accuracy and efficiency of ML models. Feature extraction typically consists in identifying and selecting the most useful characteristics from the original dataset to transform it into a smaller and more manageable set. This is achieved through a combination of statistical and mathematical methods to extract relevant patterns and relationships between features [1]. In general, a dataset contains a large number of attributes, and not all of them are relevant for effective classification, so it is crucial to identify and select the most relevant and informative ones [2]. Utilising attribute selection and feature extraction techniques in conjunction with classification algorithms such as decision trees, random forests, support vector machines and neural networks can improve the model's performance by removing redundant information and reducing complexity. This approach not only enhances the final accuracy and minimizes the risk of overfitting but may also introduce noise and lead to longer training times [3].

Data augmentation is commonly used to increase the size and diversity of a training dataset by applying various transformations to the existing samples [4]. This process is also implemented to improve the performance of ML models by providing them with more varied and representative data to learn from. As a consequence, models can learn to identify more subtle differences between classes and become more resistant to noise and outliers [5]. It is particularly effective when the training sets are limited or imbalanced, as it helps to increase the diversity of the database. Some common data augmentation techniques include image flipping, rotation and cropping. Although traditional ML methods have achieved significant success in classification projects, they can struggle to achieve satisfactory performance when dealing with complex or high-dimensional data [6]. The careful selection of transformations in data augmentation is crucial to maintain data integrity, avoid overfitting and control inconsistencies.

Many new classification approaches have emerged in recent years due to advancements in artificial intelligence by combining multiple models in order to improve overall accuracy. Ensemble methods leverage the strengths of multiple weak classifiers to produce more reliable predictions, integrating multiple results into a consistency function to get the final result with voting schemes [7]. Additionally, they have often surpassed the performance of individual models by reducing overfitting and increasing the diversity of predictions. It is a powerful technique that has improved classification within image recognition, speech recognition and natural language processing. Ensemble learning, by and large, means using data to extract features and then combining the results to predict the class of an instance with increased confidence. The combined predictions offer a more robust and accurate classification for individual instances.

## 2. Related work

Data augmentation and ensemble learning have seen significant progress in recent years, with numerous studies exploring image classification. In this section, existing literature is reviewed, highlighting some of the most relevant publications in the field.

For instance, Oyewole et al. [8] proposed an image classification approach that uses the Eigen Colour feature extraction technique based on principal component analysis (PCA). The method involves combining the predictions of the classifiers using a weighted voting scheme to generate the final prediction. Their results show that the ensemble learning approach outperforms several other methods in terms of classification accuracy, achieving 90.67%. This indicates that the method effectively captures the distinguishing colour information of the given images, leading to improved classification performance.

The second paper, by Kumar et al. [9], presents a new ensemble approach for classifying the modality of medical images. The proposed method uses multiple fine-tuned convolutional neural networks (CNNs) as feature extractors, enabling the networks to capture diverse information from various modalities. The experimental results demonstrate that the proposed ensemble accurately classifies the majority of images and achieves higher classification scores. Shijie et al. [10] also proposed a data augmentation strategy for image classification with CNNs. Their approach involves generating augmented data by applying operations such as flipping, rotation and scaling on the original dataset. The trained model was then tested on CIFAR-10 and CIFAR-100 datasets, showing that data augmentation yielded a 4% accuracy improvement on the first dataset and a 6% on the second one, compared to models trained on the original datasets.

Finally, Aggarwal et al. [11] evaluated the use of data augmentation in machine learning image recognition of five dermatological disease manifestations. Image databases were used with and without data augmentation. The model trained with data augmentation had an increase in the area under the curve (AUC) of each of the five dermatological manifestations compared to the non-augmented model.

## 3. EXPERIMENTAL DESIGN

This paper presents a research study to analyse the effects of a new perspective on ensemble learning for image classification. The primary goal of any classification method is to accurately classify the input data based on its characteristics. This paper introduces a new approach that involves using a specific data augmentation technique in order to improve the performance of standard classification methods.

The first and core step of this paper consists in generating a testing set derived from the original one using rotational augmentation techniques. After that, ensemble-based methods have been implemented and applied to every augmented set to combine the outputs and obtain a final prediction for every original case. Generally, increasing the number of outputs for ensemble learning can lead to better performance compared to standard classification, but it also comes with increased complexity and training time.

The final assessment of the paper would depend on the improvement obtained by this new approach. If the results show an improvement in the accuracy compared to previous methods, it would suggest that the specific data augmentation technique and ensemble-based methods show promise for being further applied in image classification research. On the other hand, if the results do not show a significant enhancement; it would suggest that further investigation is necessary to identify the reasons for the lack of improvement and to explore potential avenues for future research. However, it is important to note that the assessment of a research paper should not only focus on the improvement obtained but also on the limitations and potential drawbacks.

When conducting experiments, it is important to carefully select appropriate datasets to ensure a comprehensive evaluation of the proposed approach. In this paper, three different databases have been chosen as they represent diverse content, varying in terms of the number of classes and the number of samples in each class: the Chess database, the SportBall database and the Animal database. The first of them contained images of the six pieces used in chess, with at least 61 samples in each class. The main objective of the SportBall database consisted in classifying images of balls used in different sports, including basketball, tennis and bowling. It contained nine different classes, with each class having 340 samples. Lastly, the Animal database consisted of images of three different animals: cats, dogs and wild animals, with 500 samples in each class. The number of instances per class or the diversity of the classes represented in each database was not a problem because the same number of instances was selected for each class in the preprocessing phase.

## 4. PROPOSED APPROACH

A standard classification approach implicates building a predictive model using labelled training data, where each observation has a known label. The model is trained to recognise patterns and relationships within the input data and make predictions on totally unseen sets.

In this paper, a new classification method is suggested based on ensemble learning, which involves combining multiple outputs from small subsets. Figure 1 shows an overview of the proposed approach where three main steps are followed. To begin with, a new set for testing is generated by applying rotational augmentation techniques to every image of the training set, creating a set of new images that are variations of the original ones. Next, feature extraction is applied to both the original training set and the newly generated testing set. Subsequently, diverse models are generated by training the original dataset with different classifiers, upon which the augmented sets are tested. Finally, voting schemes are applied to select the most convincing results for each original case. The resulting prediction is considered to be the most reliable, as it is based on voting techniques.

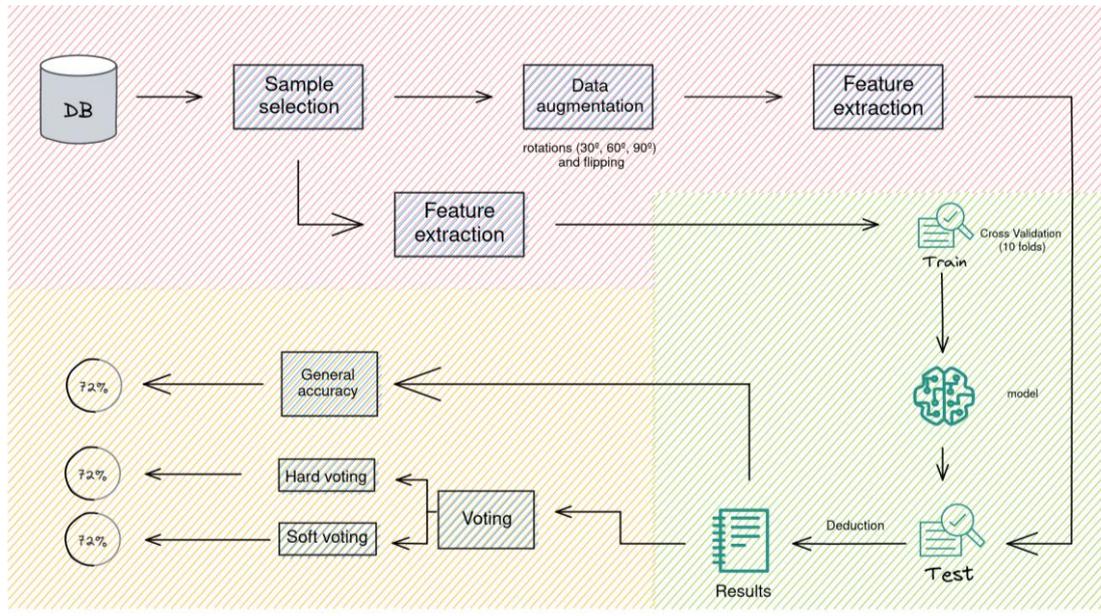

Figure 1. An overview of the proposed approach.

## 4.1. Preprocessing and feature extraction

The proposed approach is based on the utilisation of diverse data augmentation strategies on the initial training set to build the testing set. It is important to note that the databases used in this research are not always balanced, so a random sample selection has previously been applied to avoid this problem.

Sample selection involves selecting a representative subset of data from a dataset, which may be useful to reduce the computational complexity of the algorithms. In imbalanced datasets, specialised sampling techniques, such as oversampling the minority class, ensure that the model is not biased towards the majority class [12]. Generally, the number of classes is not an important point to consider in order to achieve acceptable results, but the number of samples indeed is: the more samples used to train a model, the more appropriate feature extraction is performed. As a consequence, the first step of the experiment consists in selecting an equal number of samples from each class. Initially, this value is determined by counting the samples in the smallest class. Subsequently, to analyse the effect of varying the number of samples, it is reduced to 80%, 50%, and 30% of its original value. This process results in the formation of distinct training and testing sets for each modified dataset, allowing for a comprehensive evaluation of the impact on the model's performance.

In order to generate the testing set, as shown in Figure 2, each image of the original dataset is put through transformation techniques in order to create new and modified versions: at first, every image is three times rotated in an anti-clockwise direction (30º, 60º and 90º), and the results are then flipped together with the original one. The process involves moving each pixel of the image to a new position based on the specified angle of rotation and the ones that move outside the original image dimensions are filled with the background colour. As a result, the dataset size was increased by a factor of eight, with each class containing the same number of images. For instance, if the mentioned data augmentation process is applied to a set with N samples, an eight times larger database, with 8*N samples, is obtained.

Identifying and selecting the most relevant and useful features in a dataset is essential to improve the model's performance. The Waikato Environment for Knowledge Analysis (WEKA) [13] provides various feature extraction methods, called filter methods, which preprocess datasets by selecting and removing attributes based on their statistical properties and relationships with the class variable. The extracted features are essentially numerical representations of the image characteristics that are relevant to the classification task. Examples of such features include texture, colour, shape, and intensity. However, not all of the extracted features may be equally relevant to the proposed goal, and some may be noisy or redundant, which can degrade the performance of the classification model.

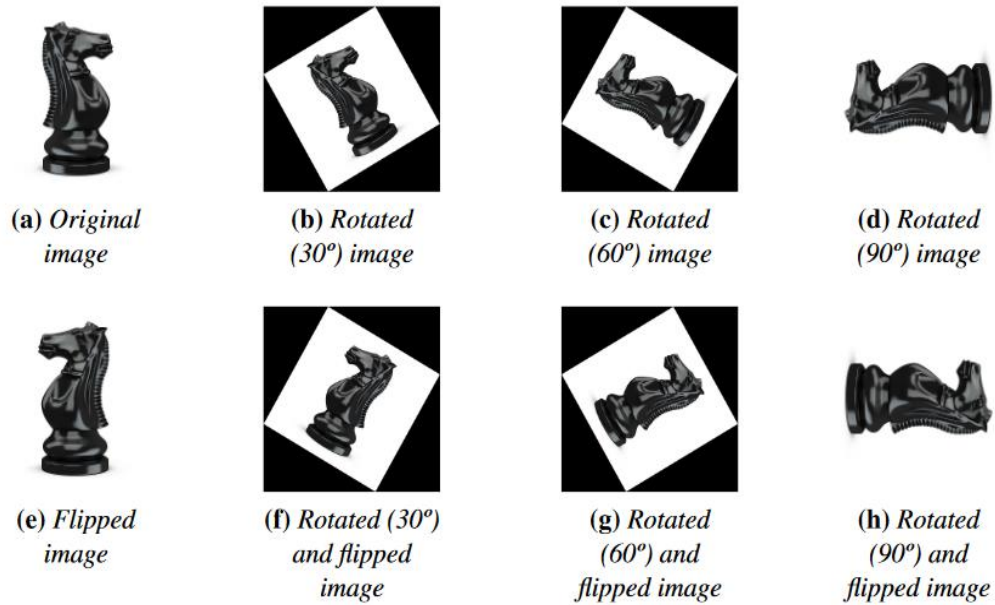

Figure 2. Set of images composed of the original instance and the augmented cases.

In this approach, the following descriptors have been applied: (i) AutoColorCorrelogramFilter (AutoColor), (ii) EdgeHistogramFilter (Edge), (iii) FuzzyOpponentHistogramFilter (Fuzzy) and (iv) PHOGFilter (PHOG). Overall, the consequence of using filter methods in WEKA is to preprocess the data in a way that improves the accuracy and performance of ML algorithms. The selection of a particular descriptor would rely on the dataset's attributes and the objective of the analysis.

(i)   The AutoColor descriptor calculates the probability of locating a pixel with a specific colour at a certain distance from another pixel with the same colour.

(ii)  The Edge descriptor computes the histogram of edge orientations in an image using the Canny edge detection algorithm.

(iii) The Fuzzy descriptor computes histograms showing how unclear each colour in the image is.

(iv)  The PHOG descriptor divides the image into smaller blocks and calculates the histogram of local edge directions and gradients.

### 4.2. Training and testing

In ML projects, the second step often involves creating a model that effectively tests data. The main task is to use the training data to generate a high-performing model that can accurately classify unknown data, which means selecting an appropriate algorithm and tuning its hyperparameters to achieve the best possible performance.

During the training phase, several models were constructed using cross-validation. In k-fold cross-validation, the dataset is divided into k subsets of equal size, where the model is trained on k-1 subsets and tested on the remaining one. To provide an overall estimate of performance, the metrics obtained from each fold were averaged, providing a more accurate analysis of the model than just testing it on a single set of data. For every training set, several ML algorithms, including (v) K-Nearest Neighbour with a value of K equal to 5 (5NN), (vi) C4.5 algorithm (J48), (vii) Naive Bayes (NB), (viii) Random Forest (RF) and (ix) Sequential Minimal Optimization (SMO) have been used [14].

(v)   The 5NN algorithm classifies data based on the majority class of its 5 nearest neighbours.

(vi)  The J48 classifier is a decision tree based on the most informative features.

(vii) The NB classifier calculates the probability of a new instance belonging to each class.

(viii) The RF classifier is an ensemble learning algorithm that constructs multiple decision trees and combines their predictions to improve accuracy.

(ix)  The SMO classifier is an efficient algorithm that separates data into different classes by finding the hyperplane with the largest margin between the classes.

Once the model has been trained, it is ready to make predictions on new data during the testing phase, where the augmented data is used to evaluate the performance of the model. The accuracy rate for the testing set is calculated to determine whether or not the constructed model has achieved satisfactory results as a general evaluation.

### 4.3. Voting schemes

Instead of applying ensemble methods to the results obtained from different classifiers, this approach employs two voting systems, hard and soft voting, to each set acquired from the data augmentation phase. Analysing every eight augmented instances and computing the most representative prediction for the entire set, a final result is obtained for each original case. This comprehensive analysis enables a thorough evaluation of the data, considering the variability introduced by the augmented instances.

Regarding the schemes, the hard voting process involves taking a majority vote among the predictions of the augmented sets. In other words, the class that receives the highest number of votes is selected as the final prediction for that particular subset [15]. Table 1 displays a randomly selected subset, presenting the predictions for the eight instances: each prediction corresponds to different categories, represented by the values C1, C2, and C3. The final prediction is determined by selecting the class with the maximum number of outputs, which in this case is Class C2 with four predictions.

Table 1. Predictions of a specific augmented set.

| I1 | I2 | I3 | I4 | I5 | I6 | I7 | I8 |
|----|----|----|----|----|----|----|----|
| C2 | C2 | C1 | C1 | C2 | C3 | C1 | C2 |

In contrast, soft voting is a more refined approach that takes into account the probability distribution generated by the model. Rather than simply counting the number of votes, this voting scheme computes the average probability of each class across all the instances [16]. Therefore, the final prediction is then made based on the class with the highest average probability. For example, as shown in Table 2, the model expresses a 62% level of confidence that I2 belongs to C1, 25% confidence that it belongs to C2, and 13% confidence that it belongs to C3. Considering the average probability, the model shows a 64% confidence level that the entire set belongs to C1.

Table 2. Probability distribution of a specific augmented set.

| I1 | I2 | I3 | I4 | I5 | I6 | I7 | I8 |
|---|---|---|---|---|---|---|---|
| [0.8, 0.1, 0.1] | [0.62, 0.25, 0.13] | [0.15, 0.8, 0.05] | [0.3, 0.05, 0.65] | [0.7 0.15, 0.15] | [0.05, 0.05, 0.9] | [0.3, 0.4, 0.3] | [0.95, 0.02, 0.03] |

[**0.64**, 0.24, 0.1]

## 5. EXPERIMENTAL RESULTS

The tables and images presented in the section illustrate the global results of the study, providing a comprehensive overview of the findings. On the one hand, the tables provide detailed information on the performance of each classifier on different descriptors [17]. Moreover, each table highlights the best sample size according to the sum of improvements in both hard and soft voting systems. The tables also offers the results obtained by standard classification methods. On the other hand, the figures visually represent the highlighted results in a colour-coded format, with grey representing the normal accuracy, blue representing the soft voting system and orange representing the hard voting system.

The results shown in Table 3 correspond to the Chess database, where the RF classifier performs the best across the Fuzzy descriptor in soft voting accuracy, ranging from 0.73 to 0.8, and across Edge, Fuzzy and PHOG descriptors in hard voting, ranging from 0.78 to 0.88. Additionally, it is mentioned that the accuracy of simple voting methods is generally enhanced by both hard voting and soft voting ensembles, which can be observed in Figure 3.

Table 3. Results for the Chess database.

| Descriptor | Classifier | Samples | Accuracy | Soft voting | Hard voting |
|---|---|---|---|---|---|
| AutoColor | 5NN | 19 | 0.31 | 0.35 | 0.37 |
| | | 31 | 0.3 | 0.38 | 0.39 |
| | | 49 | 0.29 | 0.37 | 0.33 |
| | | 61 | 0.29 | 0.37 | 0.36 |
| | J48 | 19 | 0.34 | 0.43 | 0.4 |
| | | 31 | 0.33 | 0.42 | 0.47 |
| | | 49 | 0.31 | 0.44 | 0.48 |
| | | 61 | 0.3 | 0.38 | 0.38 |
| | NB | 19 | 0.33 | 0.42 | 0.42 |
| | | 31 | 0.29 | 0.37 | 0.37 |
| | | 49 | 0.27 | 0.31 | 0.32 |
| | | 61 | 0.26 | 0.29 | 0.3 |
| | RF | 19 | 0.43 | 0.61 | 0.78 |
| | | 31 | 0.41 | 0.59 | 0.73 |
| | | 49 | 0.39 | 0.6 | 0.78 |
| | | 61 | 0.39 | 0.58 | 0.78 |
| | SMO | 19 | 0.43 | 0.61 | 0.47 |
| | | 31 | 0.4 | 0.56 | 0.54 |
| | | 49 | 0.39 | 0.54 | 0.5 |
| | | 61 | 0.39 | 0.6 | 0.48 |
| Edge | 5NN | 19 | 0.31 | 0.37 | 0.45 |
| | | 31 | 0.29 | 0.33 | 0.41 |
| | | 49 | 0.3 | 0.39 | 0.44 |
| | | 61 | 0.31 | 0.44 | 0.44 |
| | J48 | 19 | 0.36 | 0.53 | 0.63 |
| | | 31 | 0.36 | 0.46 | 0.49 |
| | | 49 | 0.35 | 0.52 | 0.57 |
| | | 61 | 0.35 | 0.52 | 0.54 |
|  | NB | 19 | 0.27 | 0.22 | 0.25 |
|  |  | 31 | 0.26 | 0.24 | 0.24 |
|  |  | 49 | 0.23 | 0.23 | 0.25 |
|  |  | 61 | 0.24 | 0.22 | 0.23 |
|  | RF | 19 | 0.42 | 0.61 | 0.82 |
|  |  | 31 | 0.41 | 0.6 | 0.8 |
|  |  | 49 | 0.41 | 0.61 | 0.86 |
|  |  | 61 | 0.4 | 0.6 | 0.86 |
|  | SMO | 19 | 0.43 | 0.6 | 0.55 |
|  |  | 31 | 0.4 | 0.55 | 0.46 |
|  |  | 49 | 0.36 | 0.45 | 0.46 |
|  |  | 61 | 0.37 | 0.5 | 0.47 |
| Fuzzy | 5NN | 19 | 0.31 | 0.33 | 0.38 |
|  |  | 31 | 0.31 | 0.38 | 0.4 |
|  |  | 49 | 0.29 | 0.36 | 0.31 |
|  |  | 61 | 0.3 | 0.32 | 0.39 |
|  | J48 | 19 | 0.47 | 0.56 | 0.51 |
|  |  | 31 | 0.45 | 0.45 | 0.49 |
|  |  | 49 | 0.46 | 0.57 | 0.57 |
|  |  | 61 | 0.45 | 0.53 | 0.53 |
|  | NB | 19 | 0.23 | 0.3 | 0.29 |
|  |  | 31 | 0.24 | 0.27 | 0.25 |
|  |  | 49 | 0.19 | 0.21 | 0.2 |
|  |  | 61 | 0.19 | 0.23 | 0.23 |
|  | RF | 19 | 0.59 | 0.8 | 0.88 |
|  |  | 31 | 0.6 | 0.75 | 0.81 |
|  |  | 49 | 0.57 | 0.74 | 0.78 |
|  |  | 61 | 0.57 | 0.73 | 0.81 |
|  | SMO | 19 | 0.26 | 0.36 | 0.26 |
|  |  | 31 | 0.26 | 0.29 | 0.27 |
|  |  | 49 | 0.21 | 0.26 | 0.3 |
|  |  | 61 | 0.23 | 0.29 | 0.27 |
| PHOG | 5NN | 19 | 0.3 | 0.38 | 0.39 |
|  |  | 31 | 0.31 | 0.34 | 0.41 |
|  |  | 49 | 0.32 | 0.45 | 0.51 |
|  |  | 61 | 0.32 | 0.41 | 0.47 |
|  | J48 | 19 | 0.36 | 0.51 | 0.51 |
|  |  | 31 | 0.34 | 0.46 | 0.5 |
|  |  | 49 | 0.33 | 0.36 | 0.39 |
|  |  | 61 | 0.35 | 0.58 | 0.59 |
|  | NB | 19 | 0.29 | 0.29 | 0.42 |
|  |  | 31 | 0.27 | 0.25 | 0.43 |
|  |  | 49 | 0.25 | 0.22 | 0.41 |
|  |  | 61 | 0.25 | 0.25 | 0.41 |
|  | RF | 19 | 0.44 | 0.65 | 0.81 |
|  |  | 31 | 0.41 | 0.61 | 0.82 |
|  |  | 49 | 0.4 | 0.63 | 0.81 |
|  |  | 61 | 0.4 | 0.65 | 0.86 |
|  | SMO | 19 | 0.46 | 0.62 | 0.52 |
|  |  | 31 | 0.42 | 0.51 | 0.46 |
|  |  | 49 | 0.42 | 0.57 | 0.55 |
|  |  | 61 | 0.43 | 0.77 | 0.58 |

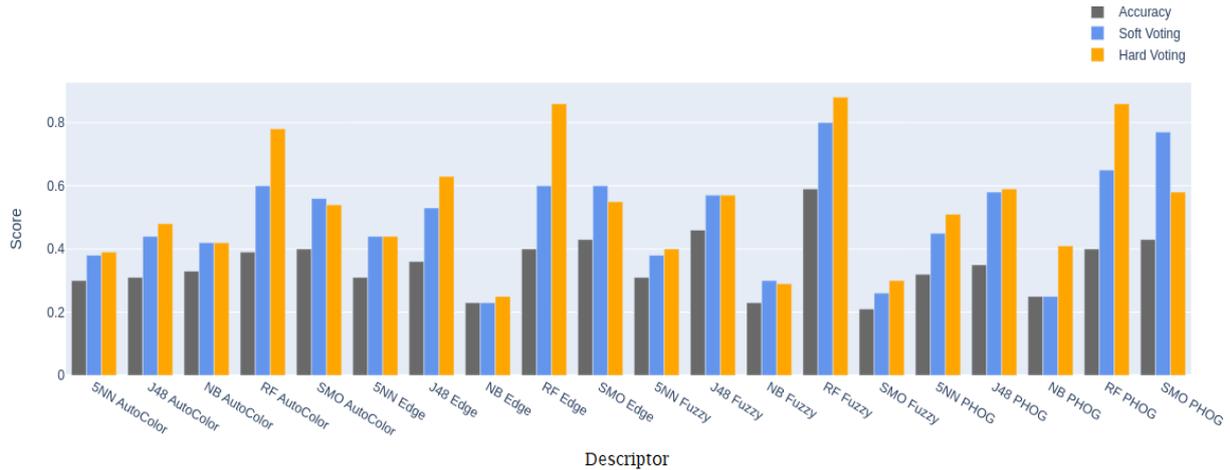

Figure 3. Filtered results for the Chess database.

The results shown in Table 4 correspond to the SportBall database, where the RF classifier also performs the best across the Fuzzy descriptor, ranging from 0.82 to 0.83 in soft voting accuracy and from 0.91 to 0.93 in hard voting. Although the general scores are much lower than in the previous case, it is worth noting that both voting ensembles, which can be observed in Figure 4, generally enhance the accuracy of simple voting methods. In certain cases, like the result achieved by the NB classifier on the Fuzzy descriptor, the standard model demonstrates superior performance compared to the hard voting system.

Table 4. Results for the SportBall database.

| Descriptor | Classifier | Samples | Accuracy | Soft voting | Hard voting |
|---|---|---|---|---|---|
| AutoColor | 5NN | 102 | 0.25 | 0.37 | 0.39 |
| | | 170 | 0.26 | 0.36 | 0.42 |
| | | 272 | 0.27 | 0.36 | 0.42 |
| | | 340 | 0.27 | 0.37 | 0.43 |
| | J48 | 102 | 0.28 | 0.46 | 0.46 |
| | | 170 | 0.27 | 0.44 | 0.47 |
| | | 272 | 0.26 | 0.44 | 0.47 |
| | | 340 | 0.27 | 0.44 | 0.45 |
| | NB | 102 | 0.23 | 0.27 | 0.27 |
| | | 170 | 0.22 | 0.25 | 0.25 |
| | | 272 | 0.22 | 0.24 | 0.24 |
| | | 340 | 0.22 | 0.25 | 0.25 |
| | RF | 102 | 0.35 | 0.59 | 0.81 |
| | | 170 | 0.35 | 0.62 | 0.82 |
| | | 272 | 0.35 | 0.57 | 0.78 |
| | | 340 | 0.35 | 0.59 | 0.8 |
| | SMO | 102 | 0.36 | 0.55 | 0.48 |
| | | 170 | 0.35 | 0.55 | 0.47 |
| | | 272 | 0.34 | 0.51 | 0.47 |
| | | 340 | 0.33 | 0.48 | 0.44 |
| Edge | 5NN | 102 | 0.27 | 0.36 | 0.4 |
| | | 170 | 0.29 | 0.41 | 0.42 |
| | | 272 | 0.3 | 0.41 | 0.42 |
| | | 340 | 0.31 | 0.42 | 0.44 |
| | J48 | 102 | 0.3 | 0.54 | 0.55 |
| | | 170 | 0.31 | 0.53 | 0.55 |
| | | 272 | 0.31 | 0.52 | 0.55 |
| | | 340 | 0.31 | 0.52 | 0.54 |
| | NB | 102 | 0.18 | 0.17 | 0.16 |
| | | 170 | 0.18 | 0.18 | 0.17 |
| | | 272 | 0.18 | 0.17 | 0.17 |
| | | 340 | 0.18 | 0.19 | 0.18 |
| | RF | 102 | 0.36 | 0.59 | 0.83 |
| | | 170 | 0.36 | 0.6 | 0.86 |
| | | 272 | 0.36 | 0.62 | 0.86 |
| | | 340 | 0.36 | 0.63 | 0.86 |
| | SMO | 102 | 0.25 | 0.38 | 0.34 |
| | | 170 | 0.25 | 0.34 | 0.33 |
| | | 272 | 0.25 | 0.33 | 0.32 |
| | | 340 | 0.24 | 0.34 | 0.31 |
| Fuzzy | 5NN | 102 | 0.3 | 0.34 | 0.44 |
| | | 170 | 0.31 | 0.37 | 0.47 |
| | | 272 | 0.31 | 0.35 | 0.46 |
| | | 340 | 0.31 | 0.36 | 0.47 |
| | J48 | 102 | 0.51 | 0.61 | 0.59 |
| | | 170 | 0.48 | 0.6 | 0.61 |
| | | 272 | 0.51 | 0.63 | 0.65 |
| | | 340 | 0.49 | 0.61 | 0.63 |
| | NB | 102 | 0.17 | 0.2 | 0.16 |
| | | 170 | 0.17 | 0.19 | 0.16 |
| | | 272 | 0.16 | 0.18 | 0.15 |
| | | 340 | 0.16 | 0.17 | 0.15 |
| | RF | 102 | 0.62 | 0.83 | 0.92 |
| | | 170 | 0.59 | 0.82 | 0.93 |
| | | 272 | 0.6 | 0.82 | 0.91 |
| | | 340 | 0.59 | 0.82 | 0.92 |
| | SMO | 102 | 0.2 | 0.24 | 0.24 |
| | | 170 | 0.19 | 0.22 | 0.22 |
| | | 272 | 0.18 | 0.2 | 0.22 |
| | | 340 | 0.18 | 0.19 | 0.22 |
| PHOG | 5NN | 102 | 0.3 | 0.41 | 0.49 |
| | | 170 | 0.31 | 0.46 | 0.48 |
| | | 272 | 0.31 | 0.45 | 0.5 |
| | | 340 | 0.3 | 0.43 | 0.48 |
| | J48 | 102 | 0.28 | 0.46 | 0.46 |
| | | 170 | 0.27 | 0.45 | 0.47 |
| | | 272 | 0.27 | 0.47 | 0.49 |
| | | 340 | 0.27 | 0.47 | 0.49 |
| | NB | 102 | 0.22 | 0.24 | 0.24 |
| | | 170 | 0.2 | 0.22 | 0.22 |
| | | 272 | 0.2 | 0.22 | 0.21 |
| | | 340 | 0.2 | 0.21 | 0.21 |
| | RF | 102 | 0.34 | 0.59 | 0.82 |
| | | 170 | 0.34 | 0.59 | 0.82 |
| | | 272 | 0.34 | 0.6 | 0.83 |
| | | 340 | 0.34 | 0.57 | 0.81 |
| | SMO | 102 | 0.35 | 0.62 | 0.45 |
| | | 170 | 0.33 | 0.6 | 0.48 |
| | | 272 | 0.31 | 0.56 | 0.46 |
| | | 340 | 0.3 | 0.51 | 0.42 |

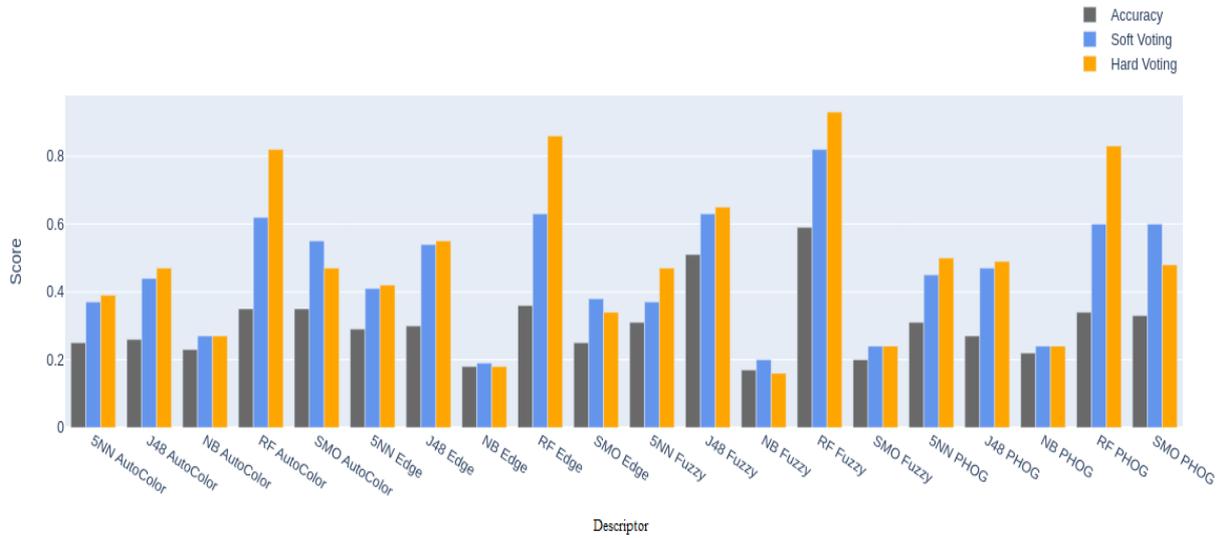

Figure 4. Filtered results for the SportBall database.

Table 5 and Figure 5 show the results for the Animal database. In general, the RF classifier tends to perform well across different types of voting and descriptors, achieving the highest accuracy in multiple cases. Hard voting seems to improve the accuracy of some classifiers, such as RF and J48, while soft voting seems to perform better for SMO. Additionally, some descriptors, such as Fuzzy and PHOG, tend to have higher accuracy overall compared to AutoColor and Edge.

Table 5. Results for the Animal database.

| Descriptor | Classifier | Samples | Accuracy | Soft voting | Hard voting |
|---|---|---|---|---|---|
| AutoColor | 5NN | 150 | 0.58 | 0.68 | 0.68 |
| | | 250 | 0.58 | 0.66 | 0.65 |
| | | 400 | 0.59 | 0.67 | 0.67 |
| | | 500 | 0.59 | 0.67 | 0.69 |
| | J48 | 150 | 0.54 | 0.71 | 0.72 |
| | | 250 | 0.56 | 0.76 | 0.75 |
| | | 400 | 0.53 | 0.72 | 0.72 |
| | | 500 | 0.54 | 0.74 | 0.76 |
| | NB | 150 | 0.49 | 0.55 | 0.56 |
| | | 250 | 0.5 | 0.54 | 0.55 |
| | | 400 | 0.5 | 0.56 | 0.57 |
| | | 500 | 0.5 | 0.57 | 0.57 |
| | RF | 150 | 0.62 | 0.8 | 0.89 |
| | | 250 | 0.61 | 0.81 | 0.87 |
| | | 400 | 0.61 | 0.83 | 0.9 |
| | | 500 | 0.62 | 0.81 | 0.88 |
| | SMO | 150 | 0.63 | 0.87 | 0.84 |
| | | 250 | 0.64 | 0.84 | 0.86 |
| | | 400 | 0.62 | 0.81 | 0.84 |
| | | 500 | 0.61 | 0.8 | 0.84 |
| Edge | 5NN | 150 | 0.51 | 0.6 | 0.62 |
| | | 250 | 0.51 | 0.62 | 0.66 |
| | | 400 | 0.53 | 0.68 | 0.73 |
| | | 500 | 0.54 | 0.68 | 0.73 |
| | J48 | 150 | 0.52 | 0.66 | 0.69 |
| | | 250 | 0.54 | 0.72 | 0.72 |
| | | 400 | 0.56 | 0.8 | 0.78 |
| | | 500 | 0.53 | 0.71 | 0.73 |
|  | NB | 150 | 0.52 | 0.64 | 0.68 |
|  |  | 250 | 0.53 | 0.65 | 0.68 |
|  |  | 400 | 0.54 | 0.68 | 0.7 |
|  |  | 500 | 0.54 | 0.68 | 0.71 |
|  | RF | 150 | 0.55 | 0.74 | 0.88 |
|  |  | 250 | 0.56 | 0.76 | 0.91 |
|  |  | 400 | 0.58 | 0.78 | 0.88 |
|  |  | 500 | 0.58 | 0.78 | 0.89 |
|  | SMO | 150 | 0.56 | 0.71 | 0.7 |
|  |  | 250 | 0.57 | 0.72 | 0.73 |
|  |  | 400 | 0.58 | 0.77 | 0.78 |
|  |  | 500 | 0.6 | 0.76 | 0.75 |
| Fuzzy | 5NN | 150 | 0.49 | 0.58 | 0.64 |
|  |  | 250 | 0.51 | 0.6 | 0.69 |
|  |  | 400 | 0.5 | 0.61 | 0.58 |
|  |  | 500 | 0.5 | 0.6 | 0.6 |
|  | J48 | 150 | 0.58 | 0.7 | 0.79 |
|  |  | 250 | 0.61 | 0.72 | 0.83 |
|  |  | 400 | 0.64 | 0.83 | 0.84 |
|  |  | 500 | 0.63 | 0.76 | 0.79 |
|  | NB | 150 | 0.41 | 0.44 | 0.44 |
|  |  | 250 | 0.39 | 0.43 | 0.41 |
|  |  | 400 | 0.39 | 0.43 | 0.41 |
|  |  | 500 | 0.4 | 0.43 | 0.42 |
|  | RF | 150 | 0.69 | 0.91 | 0.9 |
|  |  | 250 | 0.7 | 0.91 | 0.89 |
|  |  | 400 | 0.7 | 0.92 | 0.92 |
|  |  | 500 | 0.7 | 0.91 | 0.92 |
|  | SMO | 150 | 0.43 | 0.47 | 0.44 |
|  |  | 250 | 0.42 | 0.47 | 0.43 |
|  |  | 400 | 0.42 | 0.47 | 0.44 |
|  |  | 500 | 0.42 | 0.45 | 0.42 |
| PHOG | 5NN | 150 | 0.59 | 0.76 | 0.76 |
|  |  | 250 | 0.59 | 0.74 | 0.78 |
|  |  | 400 | 0.61 | 0.76 | 0.79 |
|  |  | 500 | 0.6 | 0.75 | 0.79 |
|  | J48 | 150 | 0.52 | 0.67 | 0.67 |
|  |  | 250 | 0.53 | 0.63 | 0.62 |
|  |  | 400 | 0.56 | 0.73 | 0.74 |
|  |  | 500 | 0.56 | 0.7 | 0.69 |
|  | NB | 150 | 0.48 | 0.54 | 0.6 |
|  |  | 250 | 0.47 | 0.49 | 0.54 |
|  |  | 400 | 0.48 | 0.51 | 0.55 |
|  |  | 500 | 0.48 | 0.5 | 0.56 |
|  | RF | 150 | 0.58 | 0.63 | 0.76 |
|  |  | 250 | 0.59 | 0.65 | 0.79 |
|  |  | 400 | 0.6 | 0.68 | 0.8 |
|  |  | 500 | 0.59 | 0.67 | 0.79 |
|  | SMO | 150 | 0.58 | 0.61 | 0.56 |
|  |  | 250 | 0.56 | 0.67 | 0.62 |
|  |  | 400 | 0.57 | 0.66 | 0.6 |
|  |  | 500 | 0.57 | 0.63 | 0.58 |

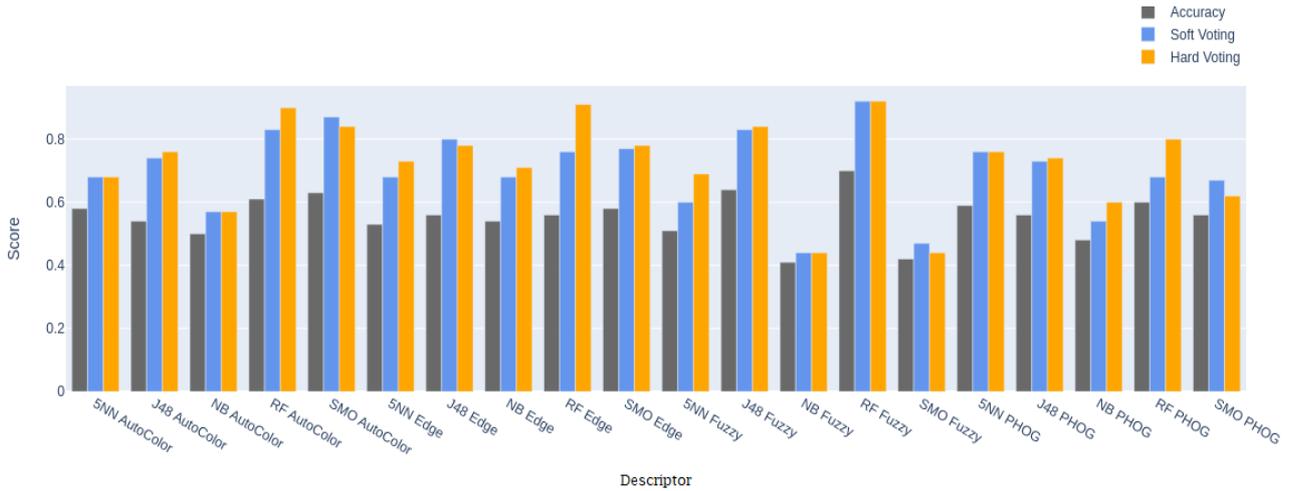

Figure 5. Filtered results for the Animal database.

Based on the previous results, it can be concluded that the RF classifier generally performs the best across all databases, with the highest accuracy values ranging above 0.8. In most cases, hard voting outperforms soft voting for most classifiers. It is also notable that NB and SMO consistently have lower accuracy than other classifiers. Finally, it can be inferred that the choice of voting method can significantly impact the accuracy of the classification models.

## 6. FUTURE WORK

Although the proposed approach has successfully achieved the main goal of the experiment, there is a need for further research in this area. For future work, it is recommended to develop a new approach based on the findings of the present study to obtain even more promising results.

In a first approximation, it is crucial to analyse the need for adjusting the number of rotations applied to each instance. By modifying the rotational angle, different testing sets can be created, increasing the number of predictions available for ensemble learning. However, it is important to note that the number of rotations applied to each instance must be carefully chosen, as too few rotations may result in limited diversity in the testing sets, while too many may introduce noise and inconsistencies. Therefore, finding the optimal number of rotations is crucial for generating effective testing sets and improving ensemble learning performance.

Data augmentation techniques such as rotational augmentation combined with voting schemes involve considering the costs and benefits that this process supposes. Therefore, it is important to strike a balance between the improvement and the additional computational resources required to implement it. However, when using data augmentation, some rotations may not be as informative as others. This is where voting schemes can be useful, as they allow for the selection of the best rotations for each class. Instead of selecting the same set of rotations for every dataset, a previous study to set these values is proposed to be applied in the future. Consequently, the rotation would depend on the category that has to be classified.

Finally, rotation processes tend to cause some of the pixels to fall outside the image boundary, creating a problem when processing the image as some parts may be missing. One way to address the current issue consists in filling the space with a background colour by extending the canvas of the image to accommodate the rotated image. Another approach that is proposed for future work is based on using interpolation techniques to fill in the missing pixels, where the values of missing data points depend on nearby values.

## CONCLUSIONS

In ML tasks, the main goal is to build a model that can generalise effectively to unseen data. It is highly important to reduce the dependencies between the training and testing sets, which means that the model should not rely too heavily on specific patterns that may not be present in the testing set.

The approach of applying voting schemes to every set of eight images is a promising new technique that has improved the precision of image analysis. In general, both hard and soft voting-based ensemble learning has led to an improvement in the accuracy of 10%-20%. While there have been cases where standard classification methods have achieved higher results, there have been also scenarios where soft voting has resulted in a 50% improvement. However, it is essential to note that this approach may introduce some degree of dependency between the two sets, which depends on the quality and diversity of the images, the specific voting scheme, and the size of the overall dataset. Therefore, it is crucial to evaluate the performance of this approach carefully and compare it to other methods to determine its effectiveness and limitations. On the whole, the use of voting schemes for image analysis shows great potential for achieving more accurate results.

Improving the accuracy of predictions has led to reducing the impact of individual model biases and errors, achieving more reliable predictions and a more generalisable model. While voting schemes can provide several benefits, there are also some potential drawbacks to using this approach: implementing ensemble learning can add additional complexity to a ML system, making it more challenging to interpret the results and understand which instance contributed to a particular prediction.

It is important to ensure that the data used to train the model is representative of the real-world scenario the model will be used in. If the data is imbalanced, meaning that one class is significantly more represented, the model may learn to predict the majority class more accurately, while neglecting the minority class. Balancing the classes gives the model an equal opportunity to learn from each category, enabling it to make more precise predictions. Creating multiple datasets with varying sample sizes has helped to improve the average accuracy rate of the voting schemes. This approach helps to identify the optimal number of samples required for acquiring desirable performance levels. Additionally, it provides insights into how the model performs under different sample size conditions.

By continuing to explore and refine these techniques, we can lead to new insights and applications in a wide range of domains.

## ACKNOWLEDGEMENTS

This work has been partially funded by the Basque Government, Spain, under Grant number IT1427-22; the Spanish Ministry of Science (MCIU), the State Research Agency (AEI), the European Regional Development Fund (FEDER), under Grant numbers TED2021-129488B-I00 and PID2021-122402OB-C21 (MCIU/AEI/FEDER, UE).

## REFERENCES

[1] S. Gnanambal, M. Thangaraj, V. T. Meenatchi, and V. Gayathri, "Classification Algorithms with Attribute Selection: an evaluation study using WEKA," vol. 9, no. 6, pp. 3640–3644, May 2018, [Online]. Available: https://search.proquest.com/docview/2059600908.

[2] N. E. rasan and D. K. Mani, "A Survey on Feature Extraction Techniques," vol. 3, no. 1, pp. 52–55, Feb. 2015, doi: 10.15680/ijircce.2015.0301009.

[3] G. Kumar and P. K. Bhatia, "A Detailed Review of Feature Extraction in Image Processing Systems," Feb. 2014, pp. 5–12, doi: 10.1109/ACCT.2014.74.

[4] C. G. Simhadri and H. K. Kondaveeti, "Automatic Recognition of Rice Leaf Diseases Using Transfer Learning," Agronomy, vol. 13, no. 4, p. 961, Mar. 2023, doi: 10.3390/agronomy13040961.

[5] M. D Bloice, C. Stocker, and A. Holzinger, "Augmentor: An Image Augmentation Library for Machine Learning," vol. 2, no. 19, p. 432, Nov. 2017, doi: 10.21105/joss.00432.

[6] X. DONG, Z. YU, W. CAO, Y. SHI, and Q. MA, "A survey on ensemble learning," vol. 14, no. 2, pp. 241–258, Apr. 2020, doi: 10.1007/s11704-019-8208-z.

[7] Z.-H. Zhou, Ensemble Methods Foundations and algorithms. Boca Raton, FL: Taylor & Francis, 2012.

[8] S. A. Oyewole and O. O. Olugbara, "Product image classification using Eigen Colour feature with ensemble machine learning," vol. 19, no. 2, pp. 83–100, Jul. 2018, doi: 10.1016/j.eij.2017.10.002.

[9] A. Kumar, J. Kim, D. Lyndon, M. Fulham, and D. Feng, "An Ensemble of Fine-Tuned Convolutional Neural Networks for Medical Image Classification," vol. 21, no. 1, pp. 31–40, Jan. 2017, doi: 10.1109/JBHI.2016.2635663.

[10] Jia Shijie, Wang Ping, Jia Peiyi, and Hu Siping, "Research on data augmentation for image classification based on convolution neural networks," Oct. 2017, pp. 4165–4170, doi: 10.1109/CAC.2017.8243510.

[11] 1st Lt. Pushkar Aggarwal, "Data augmentation in dermatology image recognition using machine learning," vol. 25, no. 6, pp. 815–820, Nov. 2019, doi: 10.1111/srt.12726.

[12] Ramyachitra, D., & Manikandan, P. (2014). Imbalanced dataset classification and solutions: a review. International Journal of Computing and Business Research (IJCBR), 5(4), 1-29.

[13] Eibe Frank, Mark A. Hall, and Ian H. Witten (2016). The WEKA Workbench. Online Appendix for "Data Mining: Practical Machine Learning Tools and Techniques", Morgan Kaufmann, Fourth Edition, 2016.

[14] M. A. U. H. Tahir, S. Asghar, A. Zafar, and S. Gillani, "A Hybrid Model to Detect Phishing-Sites Using Supervised Learning Algorithms," Dec. 2016, pp. 1126–1133, doi: 10.1109/CSCI.2016.0214.

[15] A. S. Assiri, S. Nazir, and S. A. Velastin, "Breast tumor classification using an ensemble machine learning method," vol. 6, no. 6, p. 39, May 2020, doi: 10.3390/JIMAGING6060039.

[16] A. Nayyar Hassan and A. El-Hag, "Two-Layer Ensemble-Based Soft Voting Classifier for Transformer Oil Interfacial Tension Prediction," vol. 13, no. 7, p. 1735, Apr. 2020, doi: 10.3390/en13071735.

[17] Garner, S. R. (1995, April). Weka: The waikato environment for knowledge analysis. In Proceedings of the New Zealand computer science research students conference (Vol. 1995, pp. 57-64).

## AUTHORS

Unai is a PhD student with a Bachelor's Degree in Informatics Engineering and a Master in Computational Engineering and Intelligent Systems from the University of the Basque Country. His research interests include machine learning, medical image analysis and deep learning applications.

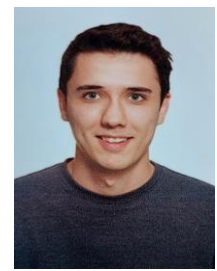